\documentclass{article}
\usepackage[final]{colm2026_conference}

\usepackage{microtype}
\usepackage{hyperref}
\usepackage{url}
\usepackage{booktabs}
\usepackage{graphicx}
\usepackage{float}
\usepackage{fvextra}
\usepackage{caption}

\definecolor{darkblue}{rgb}{0, 0, 0.5}
\hypersetup{colorlinks=true, citecolor=darkblue, linkcolor=darkblue, urlcolor=darkblue}

\title{Conversation as Measurement in Clinical Encounters: \\Observable Phase Structure, \\Partially Observable Patient State}

\author{Lily Chen, Ted Mau, Michael Gensheimer, Brian Anthony Nuyen,\\{\bfseries Nancy Jiang\thanks{Equal advising.} ~\& James Zou\footnotemark[2]}\\
Stanford University\\
\texttt{\{l1ly, tedmau, mgens, banuyen, njiang2, jamesz\}@stanford.edu}}

\begin{document}

\setcounter{footnote}{1}
\maketitle

\begin{abstract}
Many modern AI systems analyze conversational traces to infer aspects of human interaction and state, implicitly assuming that such information is recoverable from conversation. We study observability: whether a target is recoverable from conversational transcripts alone. Observability is difficult to assess because transcripts may provide only a partial view of many targets, and large-scale analysis requires model-based annotation, making true limits of the conversational signal hard to distinguish from annotator error. We therefore study clinical encounters, where patient-reported outcome measures (PROMs) provide an external anchor for patient state, and visits follow broadly structured patterns. We study observability of patient state and conversational phase structure using 439 real-world clinical encounter transcripts spanning 134 hours, including 245 ENT transcripts paired with 273 PROM surveys. We operationalize patient state using PROM scores for voice, cough, and swallowing; phase structure using conversational phase segmentation. To make these analyses credible at scale, we use a PHI-compliant GPT-5 deployment for transcript annotation and conduct 40 hours of manual validation, reducing the risk that apparent limits of observability simply reflect annotator error. Our core finding is an observability asymmetry: phase structure is observable and useful for characterizing clinical encounter organization, while patient state is only partially observable, even in a setting designed to elicit patient symptoms and experiences, cautioning against transcript-only inference of human state. 
Code is available at \url{https://github.com/lilywchen/conversation-as-measurement}.

\end{abstract}

\section{Introduction}
Many modern AI systems analyze conversational traces to infer aspects of human interaction and state \citep{ouyang2022traininglanguagemodelsfollow, openai2024gpt4technicalreport, shuster2022blenderbot3deployedconversational, rashkin2019empatheticopendomainconversationmodels}. These systems implicitly assume that such information is recoverable from conversation. But the limits of that assumption remain unclear: what can be recovered from conversation alone? This matters because conversational transcripts are increasingly used to support downstream assessment and monitoring \citep{Malgaroli2023, doi:10.1177/00472395231178943}, even when it is unclear which targets are recoverable from the conversational signal. In this work, we use \textit{observability} to mean whether a target is recoverable from conversational transcripts alone. 

Two key challenges make observability difficult to study. First, for many targets of interest that are not inherent to the conversation, such as functional state, thoughts, or emotions, conversation provides only a partial view: relevant information may never be explicitly expressed, so the absence of evidence in the transcript is hard to interpret without some external anchor. Second, because analyzing long transcripts at scale requires model-based annotation, any apparent limit of observability may reflect either a true limit of the conversational signal or an error of the annotator, which is tricky to disentangle.

The first challenge calls for a setting with an external anchor for the target of interest. We therefore design our study around clinical encounters, which often provide such an anchor through patient-reported outcome measures (PROMs) \citep{Deshpande2011, Black2013}. Clinical encounters also provide reason to expect recurring conversational structure, since clinicians are trained to conduct visits in broadly structured ways. We leverage these two properties to study observability in clinical encounters along two dimensions: patient state and conversational phase structure. These dimensions let us contrast a target that may be only partially expressed in conversation with one that is more directly tied to the transcript itself.

We study observability along these two dimensions in 439 real-world clinical encounter transcripts spanning 134 hours, including 245 ENT transcripts paired with 273 PROM surveys. We operationalize patient state using PROM scores for voice, cough, and swallowing; phase structure using conversational phase segmentation. To make these analyses credible at scale, we use a PHI-compliant GPT-5 deployment for transcript annotation and conduct 40 hours of manual validation, reducing the risk that apparent limits of observability simply reflect annotator error.

\textbf{Core insight.} Conversation in clinical encounters exhibits an observability asymmetry: phase structure is observable, while patient state is only partially observable.

This paper makes the following contributions:
\begin{enumerate}
    \item \textbf{Conversation as Measurement.} 
We conceptualize conversation in clinical encounters as a measurement modality and introduce an observability lens for transcript-based inference.
    \item \textbf{Phase Structure is Recoverable and Useful.} We use recovered phase structure to characterize clinical encounter organization, including encounter trajectories, clinician-specific phase allocation patterns, and phase-specific question patterns.

    \item \textbf{Limits of Patient State Observability.} Using PROMs, we show that patient state is only partially observable even in a symptom-eliciting setting, offering a cautionary case for transcript-only inference of human state.

\end{enumerate}

\section{Related Work}
\paragraph{Conversation Structure and Dialogue Modeling}
Foundational work in conversation analysis and dialogue systems characterizes interaction patterns \citep{Sacks1974,Schegloff_2007,Traum1992}. Other work develops approaches for modeling conversational structure in dialogue data \citep{Core1997CodingDW,ritter-etal-2010-unsupervised,SHRIBERG2000127,stolcke-etal-2000-dialogue,galley-etal-2003-discourse}. 
We build on this work by asking whether phase structure is recoverable from conversational transcripts alone.

\paragraph{Clinical Conversation Analysis} Prior work analyzes clinician-patient interaction, including sequential organization, medical communication, and consultation phase structure \citep{Frankel1984, Maynard2005, article, Manalastas2021}. Other work uses automated methods to analyze patient-provider transcripts \citep{Wallace2013, Mayfield2014, Park2019, Rajkomar2019,enarvi-etal-2020-generating}.

\paragraph{Inferring Human Attributes and States from Language} Work in psycholinguistics links linguistic patterns to psychological traits and experiences \citep{Pennebaker2003,Tausczik2009}. 
More recent work analyzes large-scale language data to infer demographic attributes, personality, and mental health signals from text \citep{Schwartz2013,coppersmith-etal-2014-quantifying,DeChoudhury2021,Eichstaedt2015,resnik-etal-2015-beyond}. 
We revisit this line of work by asking whether patient state is observable from clinical encounter transcripts.

\paragraph{Large Language Models in Conversational Settings}
Prior work evaluates conversational abilities in LLMs as chat systems \citep{NEURIPS2023_91f18a12,bai-etal-2024-mt,ou-etal-2024-dialogbench,duan-etal-2024-botchat, touvron2023llama2openfoundation}, studies their capabilities in dialogue modeling tasks \citep{qamar-etal-2025-llms,hu-etal-2022-context,heck-etal-2023-chatgpt}, and explores their use for annotating conversational data \citep{Gilardi2023}. We instead use LLMs to probe what is observable from conversational transcripts.

\section{Methods}
\subsection{Problem Formulation: Observability from Clinical Encounter Transcripts}
We study clinical encounter transcripts as a measurement modality for two targets:

\begin{itemize}
    \item \textbf{Phase Structure:} the organization of a conversation into phases, including their sequence and relative duration.
        
    \item \textbf{Patient State:} aspects of a patient’s condition, operationalized through item-level scores from PROM surveys for voice, cough, and swallowing.

\end{itemize}
We define \textit{observability} as whether a target can be recovered from the transcript alone. We assess it differently for these two targets. 
For phase structure, we assess whether model-predicted phases are reliable under manual review. For patient state, we assess whether transcript-derived PROM scores are supported by transcript evidence under manual review and whether they align with external PROM scores.

\subsection{Dataset}
\begin{table}[h]
\centering
\small

\begin{tabular}{l c}
\toprule
\textbf{Dataset Property} & \textbf{Value} \\
\midrule
Total transcripts & 439 \\
Clinicians & 4 (3 ENT specialists, 1 radiation oncologist) \\
Speaker labels & Clinician / Patient \\
Timestamp granularity & Minute-level \\
\midrule
Average conversation duration & 18.3 minutes \\
Duration standard deviation  & 8.1 minutes \\
Shortest conversation & 4 minutes \\
Longest conversation & 51 minutes \\
\midrule
PROM surveys & 273  \\
\quad Voice & 166 \\
\quad Cough & 48 \\
\quad Swallowing & 59 \\
\bottomrule
\end{tabular}
\caption{Dataset summary. PROM surveys are available for a subset of ENT transcripts (245 transcripts paired with 273 surveys). Some encounters are paired with multiple surveys when patients have multiple concerns. 
}
\label{tab:data_summary}
\end{table}

\paragraph{Conversational Transcripts} Each encounter was ambiently recorded and transcribed using Nuance Dragon Ambient eXperience (DAX), a clinical documentation system. We analyze the DAX transcript output directly, without additional preprocessing. Transcripts include speaker labels and utterance-level timestamps.

\paragraph{PROM Surveys} We use three validated PROMs commonly used in ENT care: the Voice Handicap Index-10 (VHI-10) \citep{Rosen2004}, the Cough Severity Index (CSI) \citep{Shembel2013}, and the Eating Assessment Tool-10 (EAT-10), a measure of swallowing difficulty \citep{Belafsky2008}. Each instrument contains 10 items, with responses scored on a 0--4 ordinal scale reflecting increasing symptom burden. These surveys are administered as part of clinical care when relevant to the visit, rather than uniformly across encounters. In our setting, patients typically completed these surveys shortly before the clinical encounter.

\subsection{LLM-Based Transcript Annotation}

\paragraph{Model} We use a PHI-compliant GPT-5 deployment to segment transcripts into phases, extract questions, and infer PROM scores.

\paragraph{Phase Segmentation} We segment transcripts into phases based on conversational flow, capturing shifts in communicative purpose across the encounter (Table~\ref{tab:conversation_phases}). Segment timestamps are defined by the first and last utterances in each phase.

\renewcommand{\arraystretch}{1.25}
\begin{table}[h]
\centering
\small
\begin{tabular}{p{3.4cm} p{9.7cm}}
\toprule
\textbf{Phase} & \textbf{Definition} \\
\midrule
\textbf{Opening rapport} & Greetings, rapport-building, and initial conversation. \\

\textbf{History} & Patient symptoms, context, and patient-provided information. \\

\textbf{Physical exam} & Clinician exam instructions and observable findings; no interpretation. \\

\textbf{Assessment} & Clinician interpretation, diagnostic reasoning, or impressions. \\

\textbf{Plan recommendations} & Clinician treatment steps, tests, medications, referrals, or next steps. \\

\textbf{Education counseling} & Clinician explanations, rationale, or guidance (non-action content). \\

\textbf{Closing} & Visit wrap-up, confirming next steps, and ending the encounter. \\

\textbf{Non-clinical} & Administrative details, logistics, or unrelated small talk. \\
\bottomrule
\end{tabular}
\caption{Phase schema used to segment clinical encounter transcripts. The schema was informed by medical literature and clinician consultation.}
\label{tab:conversation_phases}
\end{table}

\paragraph{Question Extraction} 
We prompt the model to extract information-seeking questions at the utterance level and attribute each to the speaker (clinician or patient). Across 439 encounters, we identify 11,696 questions. After lowercasing, 93.3\% of extracted question strings exactly match text in the transcript. Recall was not measured.

\paragraph{Transcript-Derived PROM Scores} For encounters with paired PROM surveys, we use the model to predict transcript-derived scores for each PROM item. When no direct evidence is present in the transcript, the model is instructed to abstain from predicting a score.

\subsection{Manual Validation}
One of the authors manually examined a subset of transcripts to assess model-predicted phases, questions, and transcript-derived PROM scores.

\paragraph{Phase Validation} The reviewer examined all 212 model-predicted phases from 29 randomly sampled transcripts spanning all four clinicians.
Each phase was assessed for binary correctness by reviewing the transcript and checking two criteria: (1) correctness of the phase label and (2) consistency of the start and end timestamps with the transcript. In total, this phase validation required approximately 23 hours.

\paragraph{Question Validation} The reviewer examined 100 randomly sampled questions. Each question was assessed for binary correctness by reviewing the transcript and checking two criteria: (1) whether it was an information-seeking question that occurred in the transcript and (2) correctness of the speaker role attribution. This question validation required approximately 4 hours.

\paragraph{Transcript-Derived PROM Score Validation}
The reviewer examined all 120 items from 12 transcript--PROM survey pairs, with four pairs from each PROM instrument. Each item was assessed for binary correctness by reviewing the transcript for evidence. For scored predictions, the transcript had to contain evidence supporting the model-predicted score. For abstentions, the transcript had to lack sufficient evidence to support any score. PROM score validation required approximately 13 hours.

\subsection{Evaluation Metrics for Transcript-Derived PROM Scores}
We report three metrics for comparing transcript-derived PROM scores to corresponding external PROM scores:
\begin{itemize}
    \item Quadratic weighted kappa (QWK) \citep{Cohen1968}
    \item Mean absolute error (MAE)
    \item Mean signed error (external PROM score $-$ transcript-derived PROM score)
\end{itemize}
These metrics are computed on non-missing items where the model does not abstain.

\section{Results}
\subsection{Observability of Phase Structure}
\subsubsection{Phase Validation Results} 
\begin{table}[h]
\centering
\small
\begin{tabular}{lcc}
\toprule
\textbf{Group} & \textbf{N} & \textbf{Correct (\%)} \\
\midrule
Phases & 212 & 94.8 \\
Question Validity& 100 & 99.0\\
Question Speaker Attribution & 100 & 99.0 \\
\bottomrule
\end{tabular}
\caption{Manual validation of sampled model outputs, reporting phase correctness, question-extraction precision, and question speaker-attribution accuracy.}
\label{tab:phase_val}
\end{table}

This manual validation shows that phase structure is recoverable from clinical encounter transcripts. We next show how recovered phases characterize encounter organization at multiple levels.

\subsubsection{Overall Phase Allocation}

\begin{figure}[h]
\centering
\includegraphics[width=0.5\linewidth]{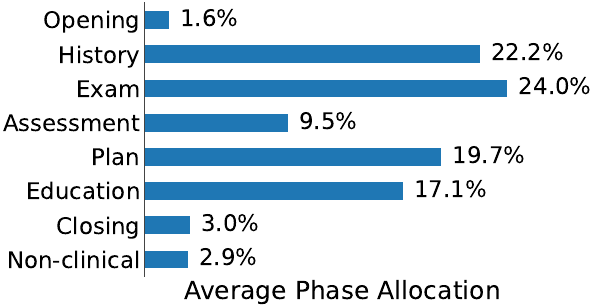}
\caption{On average, most encounter time is spent in history, exam, plan, and education, while opening, closing, and non-clinical phases occupy relatively little time.}
\caption*{\footnotesize Note: Phase allocation percentages are computed relative to total extracted segment duration; due to minute-level timestamp resolution, extracted phases cover 90\% of total conversation time on average.}
\label{fig:phase}
\end{figure}

The concentration of time in clinically substantive phases, with relatively little time devoted to rapport-oriented or off-task exchange, is consistent with the expected high-level organization of clinical encounters. As an aggregate result, this provides initial evidence that the recovered phases capture a coherent encounter structure.

\subsubsection{Encounter-Level Phase Trajectories}
\begin{figure}[h]
\centering
\includegraphics[width=\linewidth]{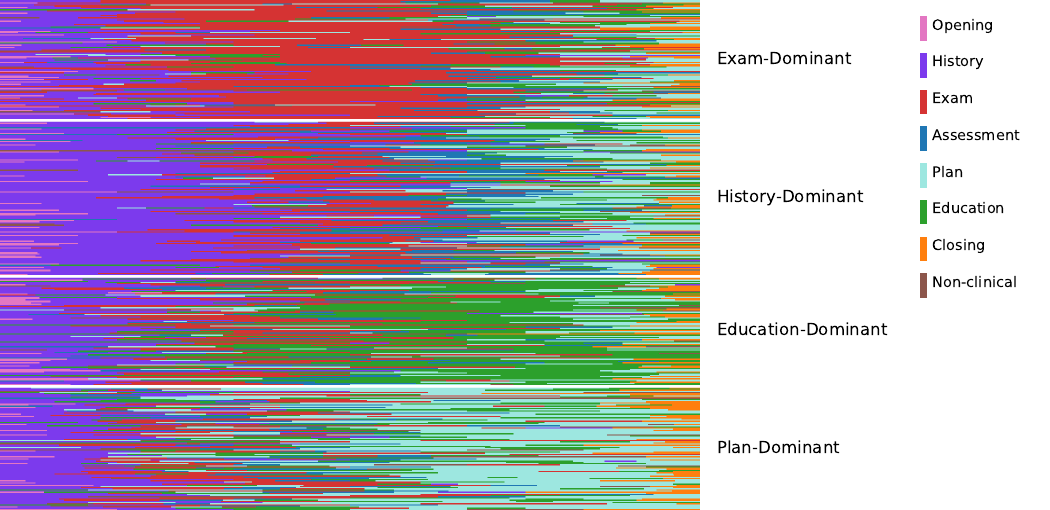}
\caption{Sequence index plot of all 439 clinical encounters, with each row showing the normalized phase trajectory of a single encounter. Encounters are grouped using K-means clustering ($K=4$) on phase allocation percentages, revealing several broad trajectory patterns; descriptive cluster labels are shown at right.}
\label{fig:sequence}
\end{figure}

Figure~\ref{fig:sequence} shows that encounter-level organization is structured in recurrent ways rather than varying arbitrarily from transcript to transcript. Across encounters, phases tend to follow a broad temporal progression, with history appearing earlier, exam often occupying substantial middle portions, and plan or education becoming more prominent later. At the same time, the plot reveals meaningful heterogeneity in how this progression is realized: some encounters devote much more of the encounter to exam, others remain history-heavy for longer, and others allocate more late-stage time to education or plan.

The resulting trajectory clusters show that recovered phase structure captures both recurring patterns and structured variation in how encounters unfold over time.
\subsubsection{Clinician-Specific Phase Allocation Patterns}
\label{sec:clinician_comm}
\begin{figure}[h]
\centering
\includegraphics[width=\linewidth]{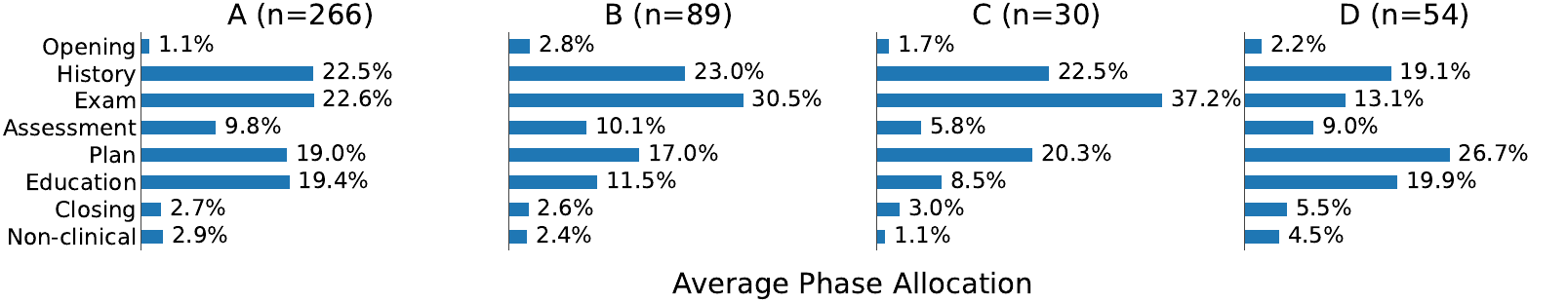}
\caption{Average phase allocation stratified by clinician. Clinicians A--C (ENT specialists) devote more encounter time to exam, while Clinician D (radiation oncology) devotes more time to plan and education.}
\label{fig:clinician}
\end{figure}

Phase allocation also differs systematically across clinicians. Among the ENT clinicians, A and B devote the largest share of encounter time to exam, with relatively similar allocations to history, while C is more exam-heavy than A and B.

More broadly, some of the variation may reflect specialty differences. Clinicians A--C are ENT specialists, where visits often involve direct examination of the throat, airway, or related structures, consistent with the larger share of encounter time spent in exam. Clinician D is in radiation oncology, where encounters may place relatively more emphasis on cancer treatment planning, follow-up, and patient counseling, consistent with greater time in plan and education. This interpretation is suggestive rather than causal. These differences show that recovered phases capture clinician-specific patterns in encounter organization.

\subsubsection{Phase-Specific Question Patterns}
\begin{figure}[t]
\centering
\includegraphics[width=\linewidth]{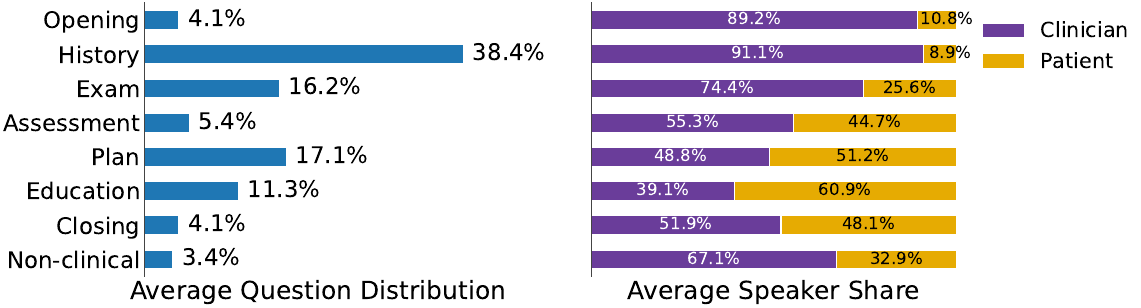}
\caption{Left: Average share of questions by phase, concentrated in history, plan, and exam.
Right: Questioning role by phase, showing the share of clinician and patient questions within each phase.}
\label{fig:questions}
\end{figure}

Questions are not distributed uniformly across phases. Figure~\ref{fig:questions} shows that the largest share of questions occurs during history, followed by plan and exam, while opening, assessment, closing, and non-clinical phases account for much smaller shares. This pattern is consistent with the functional roles of these phases: history concentrates information gathering, exam often involves clarification and elicitation, and plan includes discussion of next steps and recommendations.

Questioning is also role-dependent within phase. Clinicians ask the large majority of questions in opening, history, and exam, whereas questioning becomes more balanced in assessment and closing. In contrast, patients ask slightly more questions in plan and a clear majority in education. These phase-specific questioning patterns indicate that recovered phase structure supports analysis of how interactional roles change over the course of the encounter.

Taken together, these analyses show that recovered phase structure reveals useful insights into encounter organization at multiple levels. We next turn to patient state, where the limits of observability become more apparent.

\subsection{Partial Observability of Patient State}

\subsubsection{Manual Validation of Transcript-Derived PROM Scores}
\begin{table}[h]
\centering
\small
\begin{tabular}{lccc|ccc}
\toprule
\textbf{Group} & \textbf{N} & \textbf{Supported (\%)} & \textbf{Missing (\%)} & \textbf{QWK} & \textbf{MAE} & \textbf{Mean Signed Error}
\\
\midrule
Overall & 120 & 100.0 & 65.8 & 0.467 & 1.098 & 0.659\\
Voice & 40 & 100.0  & 67.5 & 0.606 & 0.615 & 0.0 \\
Cough & 40 & 100.0  &  75.0 & -0.019 & 1.700 & 1.700\\
Swallowing & 40 & 100.0  &  55.0 & 0.586 & 1.111 & 0.556\\
\bottomrule
\end{tabular}
\caption{Manual validation of transcript-derived PROM scores. Left: the percentage judged supported by transcript evidence and item missingness. Right: agreement with external PROM scores for non-missing items in the validation subset.}
\label{tab:val_metrics}
\end{table}

Table~\ref{tab:val_metrics} reports results for the manually reviewed subset of 120 PROM items from 12 transcript--survey pairs. All transcript-derived PROM scores and abstentions in this subset were judged to be supported by transcript evidence. For non-missing items, overall agreement with external PROM scores was only moderate (QWK 0.467; MAE 1.098), suggesting that transcripts support only partial recovery of survey responses. The cough QWK of $-0.019$ is based on only 10 non-missing items and should therefore be interpreted cautiously. These results suggest that part of the disagreement with external PROM scores may reflect limits of observability from conversation rather than model annotator error.

\subsubsection{Transcript-Derived PROM Missingness and Agreement}
\begin{table}[h]
\centering
\small
\begin{tabular}{lcc |c c c}
\toprule
\textbf{PROM Group} & \textbf{N} & \textbf{Missing (\%)} & \textbf{QWK} & \textbf{MAE} & \textbf{Mean Signed Error}\\
\midrule
Overall & 2730 & 63.0 & 0.508 & 1.006 & 0.391\\
Voice & 1660 & 65.7 & 0.488 & 1.040 & 0.430\\
Cough & 480 & 67.1 &  0.527 & 0.987 & 0.696\\
Swallowing & 590 & 52.0 & 0.539 & 0.947 & 0.134 \\
\bottomrule
\end{tabular}
\caption{Analysis of all transcript-derived PROM scores. Left: item missingness. Right: agreement metrics for non-missing items.}
\label{tab:prom_summary}
\end{table}

Table~\ref{tab:prom_summary} reports the full analysis across all 2,730 PROM items from the 273 paired PROM surveys. Transcript-derived PROM scores are frequently missing, with 63.0\% missing overall. Missingness varies across instruments, from 52.0\% for Swallowing to 67.1\% for Cough. Among non-missing items, agreement with external PROM scores is only moderate overall (QWK 0.508; MAE 1.006), with similar agreement levels across instruments. The positive mean signed error (external minus transcript-derived score; 0.391 overall and positive for all three instruments) indicates a tendency to underestimate patient-reported symptom burden among non-missing items, although this analysis cannot determine the relative contributions of limited verbal disclosure and conservative model scoring. 
Together, these results suggest that patient state is only partially observable from conversation: many PROM items are not expressed in the transcript, and even when evidence is present, transcript-derived scores show only limited agreement with patient-reported scores.

This finding is notable because clinical encounters are specifically designed to elicit symptoms and patient experience, making them a relatively favorable setting for transcript-based inference of patient state. Even here, however, observability remains limited. This sharpens the contrast with phase structure, which was recoverable from the same conversations.

\begin{figure}[h]
\centering
\includegraphics[width=\linewidth]{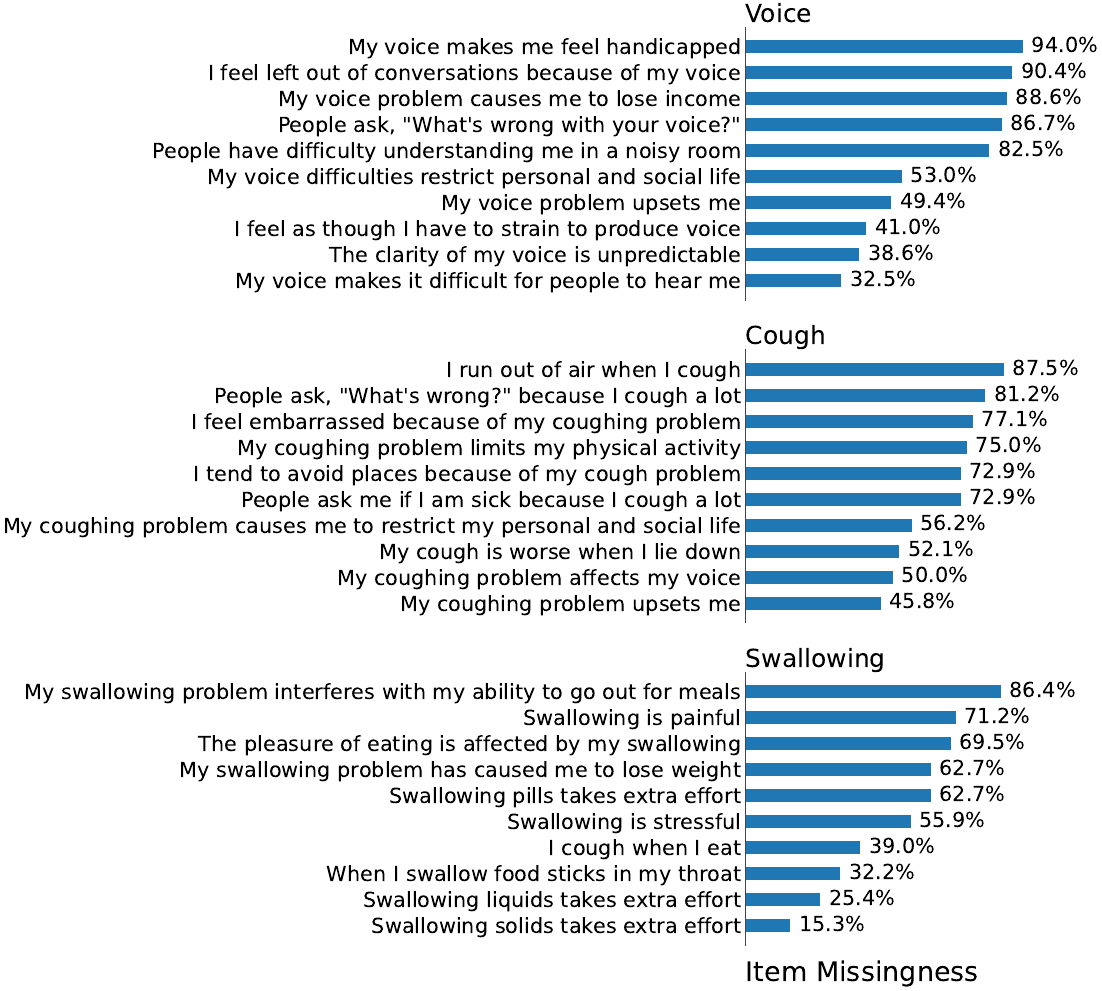}
\caption{Item missingness varies substantially within each PROM instrument, indicating that observability is item-dependent.}
\label{fig:missing_item}
\end{figure}

\subsubsection{Item-Level Missingness}
Partial observability is not uniform. Figure~\ref{fig:missing_item} shows substantial variation in missingness across items within each PROM, indicating that observability is item-dependent, not just instrument-dependent. In Voice, the most frequently missing items concern social and emotional impact, such as feeling handicapped or excluded from conversation, whereas more concrete functional items, such as difficulty being heard, are missing less often. In Cough, missingness is high across nearly all items, but some variation remains: socially mediated items, including embarrassment and others asking what is wrong, are among the most often missing, while some bodily or situational experiences, such as coughing when lying down, are somewhat less often missing. In Swallowing, items about pain, enjoyment of eating, and interference with meals outside the home are much more often missing than items about effort swallowing liquids or solids.

These patterns suggest that observability may depend in part on the kind of evidence an item requires, but not in a perfectly uniform way. In these clinical encounters, items tied to subjective, affective, or socially situated experiences often appear more likely to be absent from the transcript, whereas some concrete functional difficulties appear more likely to be verbalized. We view this as an interpretive pattern rather than a strict rule. Even within an instrument, some facets of patient state are more observable from conversation than others.

\section{Discussion}
We framed this paper around observability from conversation: whether a target is recoverable from conversational transcripts alone. Across real-world clinical encounters, we find an observability asymmetry: conversational phase structure is recoverable at scale and useful for characterizing encounter organization, whereas patient state, as anchored by PROMs, is only partially observable.

The recoverability of phase structure is consistent with the nature of the target: phase structure is expressed in how the interaction unfolds, so transcripts provide direct evidence of visit organization. This makes transcript-based structural analysis a promising use of conversational data. In our setting, recovered phase structure supported analysis of encounter trajectories, clinician-level differences in visit organization, and phase-specific question patterns, showing that transcripts can support meaningful analysis of interactional structure even without manual annotation at scale.

By contrast, patient state is only partially observable, even in a setting designed to elicit patient symptoms and experiences, because the transcript is an incomplete record of the underlying construct. Even when PROMs are administered, much of the PROM-relevant information is not verbalized in the interaction, and this missingness varies across PROM instruments and even across items within a single instrument. Among the PROM-relevant information that is expressed, agreement with patient-reported scores remains limited. Partial observability reflects a mismatch between the target and the measurement channel: some aspects of patient-reported state are available in conversation, while others are absent, underspecified, or only indirectly signaled.

This asymmetry also has a broader implication: transcript-only inference of human state, and downstream applications that rely on it, should be treated with caution unless there is good reason to believe the transcript contains sufficiently informative evidence for the target. Where that condition does not hold, external anchors such as surveys, more structured elicitation, or additional modalities may be necessary to support valid measurement.

Our study has several limitations. First, the data come from a limited clinical setting and a small set of PROM instruments, so the precise degree of observability may differ across specialties, encounter types, and measurement constructs. Other clinical settings may also follow different interactional patterns and may require refinement of the phase definitions used here. Second, although we pair LLM-based annotation with extensive manual validation, that validation was conducted by a single annotator; our conclusions still depend on model-mediated analysis and may be affected by residual annotator error. Third, gold transcripts were unavailable, so we could not assess ASR word error rate (WER); transcription errors may therefore contribute to apparent missingness. Fourth, transcripts are only one representation of the encounter and omit prosody, gesture, and other contextual cues that may also carry information about patient state. Finally, PROMs provide useful external anchors, but they are not complete ground truth for all aspects of patient experience. Together, these limitations leave open the extent to which our findings generalize across clinical settings, measurement targets, and PROM instruments.

\section*{LLM Usage Disclosure}
We used a PHI-compliant version of GPT-5 as the model annotator for phase segmentation, question extraction, and transcript-derived PROM scoring. We also used an LLM during manuscript preparation for editing support, including phrasing, flow, and figure captions, as well as for refactoring code used in figure generation and analysis.

\section*{Ethics Statement}
This study was conducted under IRB approval. The clinical transcripts contain protected health information and were processed in a secure PHI-compliant environment. We report only aggregate results and do not release raw transcripts or identifying patient information.

\section*{Reproducibility Statement}
The clinical transcripts and linked PROM data used in this study cannot be publicly released because they contain protected health information. Appendix~\ref{sec:reproducibility_appendix} provides the annotation prompts, PROM item lists, and a synthetic DAX transcript that contain no study data or protected health information. We release code for analysis, evaluation, and figure generation in the GitHub repository linked in the abstract.

\section*{Acknowledgments}
This material is based upon work supported by the U.S. Department of Energy, Office of Science, Office of Advanced Scientific Computing Research, Department of Energy Computational Science Graduate Fellowship under Award Number DE-SC0026073.

This report was prepared as an account of work sponsored by an agency of the United States Government. Neither the United States Government nor any agency thereof, nor any of their employees, makes any warranty, express or implied, or assumes any legal liability or responsibility for the accuracy, completeness, or usefulness of any information, apparatus, product, or process disclosed, or represents that its use would not infringe privately owned rights. Reference herein to any specific commercial product, process, or service by trade name, trademark, manufacturer, or otherwise does not necessarily constitute or imply its endorsement, recommendation, or favoring by the United States Government or any agency thereof. The views and opinions of authors expressed herein do not necessarily state or reflect those of the United States Government or any agency thereof.
\bibliography{colm2026_conference}
\bibliographystyle{colm2026_conference}
\newpage
\appendix
\section{Reproducibility Materials}
\label{sec:reproducibility_appendix}

The templates below show the annotation instructions used in the study. At runtime, \texttt{TRANSCRIPT} was replaced with the raw DAX transcript and \texttt{ITEMS} with the ten items from the applicable PROM instrument.

\subsection{Phase Segmentation and Question Extraction Prompt}

\textbf{System message}
\begin{Verbatim}[breaklines=true,breakanywhere=true,fontsize=\footnotesize]
You are a clinical conversation annotator. Your task is to extract semantic
segments and questions from a clinician-patient transcript.

SEGMENTATION GUIDELINES:
- Split the conversation into 3-8 semantic segments (phases).
- Create a new segment only when the communicative purpose changes. Do not
  split mid-topic. If an utterance contains mixed elements, assign it to the
  segment with the dominant communicative purpose. Do not create micro-segments
  based on individual sentences.
- Use the first and last utterances in a segment to determine segment_start and
  segment_end timestamps (HH:MM AM/PM).

ALLOWED SEGMENT TITLES:
- opening_rapport: greetings, rapport-building, and initial conversation.
- history: patient symptoms, story, context, and patient-provided information.
- physical_exam: clinician exam instructions and observable findings; no
  interpretation.
- assessment: clinician interpretation, diagnostic reasoning, impressions, or
  hypotheses.
- plan_recommendations: clinician treatment steps, tests, medications,
  referrals, or explicit next-step actions.
- education_counseling: clinician explanations, rationale, or guidance.
- closing: visit wrap-up, confirming next steps, and ending the encounter.
- non_clinical: administrative details, logistics, or unrelated small talk.

QUESTION EXTRACTION RULES:
- Extract every utterance that is a genuine information-seeking question.
- Do not treat exam commands, instructions, or directives as questions.
\end{Verbatim}

\textbf{User message and output schema}
\begin{Verbatim}[breaklines=true,breakanywhere=true,fontsize=\footnotesize]
Output one JSON object with the key "segments". Each segment must contain:
{
  "segment_title": "opening_rapport | history | physical_exam | assessment |
                     plan_recommendations | education_counseling | closing |
                     non_clinical",
  "segment_start": "HH:MM AM/PM",
  "segment_end": "HH:MM AM/PM",
  "summary": "1-2 sentence summary",
  "questions": [
    {"time": "HH:MM AM/PM",
     "speaker": "clinician | patient",
     "text": "question copied verbatim in 20 words or fewer"}
  ]
}

Transcript:
{TRANSCRIPT}

Return JSON only: {"segments": [...]}
\end{Verbatim}
\newpage
\subsection{Transcript-Derived PROM Prompt}

\textbf{System message}
\begin{Verbatim}[breaklines=true,breakanywhere=true,fontsize=\footnotesize]
You are a clinical outcomes annotator trained to extract symptom severity
ratings from clinician-patient transcripts. Infer ratings for exactly ten items
and return valid JSON matching the schema below.

Rating rubric:
0 = never; 1 = almost never; 2 = sometimes; 3 = almost always; 4 = always.

Evidence rules:
- Extract every relevant evidence mention supporting the rating.
- Each evidence quote must be copied verbatim in 20 words or fewer.
- Include speaker (patient or clinician) and timestamp (HH:MM AM/PM).
- Do not fabricate, paraphrase, or alter quotes.
- Explain the rating in 1-2 sentences.

Evidence strength:
- weak: evidence is indirect, ambiguous, minimal, or partially relevant.
- strong: at least one clear, explicit, directly relevant quote supports the
  rating.

Abstention:
If the transcript provides no information relevant to an item, set score to
null, evidence to [], evidence_strength to "none", and explain that there is
no evidence.

Per-item schema:
{
  "score": 0 | 1 | 2 | 3 | 4 | null,
  "explanation": "1-2 sentence explanation",
  "evidence": [
    {"quote": "verbatim quote in 20 words or fewer",
     "speaker": "patient | clinician",
     "timestamp": "HH:MM AM/PM"}
  ],
  "evidence_strength": "none | weak | strong"
}
Return strictly valid JSON and no additional keys or text.
\end{Verbatim}

\textbf{User message}
\begin{Verbatim}[breaklines=true,breakanywhere=true,fontsize=\footnotesize]
Analyze the transcript and output one JSON object with exactly the ten item keys
listed below, in the same order and using the schema above.

Items:
{ITEMS}

Transcript:
{TRANSCRIPT}

Return JSON only.
\end{Verbatim}

\clearpage
\subsection{PROM Items}
\label{sec:prom_items}

The annotation prompt used a shared rating rubric from 0 (never) to 4 (always) for the items below.

\subsubsection{Cough Severity Index (CSI)}
\begin{enumerate}
    \item My cough is worse when I lie down.
    \item My coughing problem causes me to restrict my personal and social life.
    \item I tend to avoid places because of my cough problem.
    \item I feel embarrassed because of my coughing problem.
    \item People ask, ``What's wrong?'' because I cough a lot.
    \item I run out of air when I cough.
    \item My coughing problem affects my voice.
    \item My coughing problem limits my physical activity.
    \item My coughing problem upsets me.
    \item People ask me if I am sick because I cough a lot.
\end{enumerate}

\subsubsection{Voice Handicap Index-10 (VHI-10)}
\begin{enumerate}
    \item My voice makes it difficult for people to hear me.
    \item People have difficulty understanding me in a noisy room.
    \item My voice difficulties restrict personal and social life.
    \item I feel left out of conversations because of my voice.
    \item My voice problem causes me to lose income.
    \item I feel as though I have to strain to produce voice.
    \item The clarity of my voice is unpredictable.
    \item My voice problem upsets me.
    \item My voice makes me feel handicapped.
    \item People ask, ``What's wrong with your voice?''
\end{enumerate}

\subsubsection{Eating Assessment Tool-10 (EAT-10)}
\begin{enumerate}
    \item My swallowing problem has caused me to lose weight.
    \item My swallowing problem interferes with my ability to go out for meals.
    \item Swallowing liquids takes extra effort.
    \item Swallowing solids takes extra effort.
    \item Swallowing pills takes extra effort.
    \item Swallowing is painful.
    \item The pleasure of eating is affected by my swallowing.
    \item When I swallow food sticks in my throat.
    \item I cough when I eat.
    \item Swallowing is stressful.
\end{enumerate}

\clearpage
\subsection{Synthetic DAX Transcript}
\label{sec:synthetic_examples}

\begin{Verbatim}[breaklines=true,breakanywhere=true,fontsize=\footnotesize]
DAX Copilot
Patient session
Transcript
Recording started: Monday, January 12, 9:00 AM

Clinician 9:00 AM
What brings you in today?

Patient 9:00 AM
My voice becomes tired by the end of the workday.

Clinician 9:04 AM
I am going to examine your throat now.

Clinician 9:08 AM
The exam is reassuring and suggests vocal strain.

Clinician 9:09 AM
Let us discuss voice therapy.

Clinician 9:12 AM
Does swallowing take extra effort?

Patient 9:12 AM
Solid food sometimes feels stuck, but liquids are fine.
\end{Verbatim}

\end{document}